\def\year{2021}\relax
\documentclass[letterpaper]{article} 
\usepackage[switch]{lineno}
\usepackage{aaai21}  
\usepackage{times}  
\usepackage{helvet} 
\usepackage{courier}  
\usepackage[hyphens]{url}  
\usepackage{graphicx} 
\usepackage{natbib}  
\usepackage{caption} 
\usepackage[table,xcdraw]{xcolor}
\usepackage{amsmath}

\newcommand{\toolname}{{\bf GDSolver} }
\newcommand{\R}{\mathbb{R}}
\usepackage{amsfonts}

\usepackage{algorithm}
\usepackage{algorithmic}

\title{A Solver + Gradient Descent Training Algorithm for Deep Neural Networks}
\author{
Dhananjay Ashok$^1$
\and
Vineel Nagisetty$^2$\and
Christopher Srinivasa$^{2}$\And
Vijay Ganesh$^3$
\affiliations{
$^1$University of Toronto,\\ 
$^2$Borealis AI,\\ 
$^3$ University of Waterloo}
\emails{
dhananjay.ashok@mail.utoronto.ca\\ \{vineel.nagisetty, christopher.srinivasa\}@borealisai.com\\
vijay.ganesh@uwaterloo.ca
}
}
\author{
    Dhananjay Ashok\textsuperscript{\rm 1},
    Vineel Nagisetty\textsuperscript{\rm 3},
    Christopher Srinivasa\textsuperscript{\rm 3},
    and Vijay Ganesh\textsuperscript{\rm 2}
    \\
}
\affiliations{

    \textsuperscript{\rm 1}University of Toronto, Canada\\
    
    \textsuperscript{\rm 2}University of Waterloo, Canada\\
    \textsuperscript{\rm 3}Borealis AI, Canada\\
    dhananjay.ashok@mail.utoronto.ca,vijay.ganesh@uwaterloo.ca, \{vineel.nagisetty, christopher.srinivasa\}@borealis.ai
}

\begin{document}

\maketitle

\begin{abstract}
We present a novel hybrid algorithm for training Deep Neural Networks that combines the state-of-the-art Gradient Descent (GD) method with a Mixed Integer Linear Programming (MILP) solver, outperforming GD and variants in terms of accuracy, as well as resource and data efficiency for both regression and classification tasks. Our GD+Solver hybrid algorithm, called \toolname, works as follows: given a DNN $D$ as input, \toolname invokes GD to partially train $D$ until it gets stuck in a local minima, at which point \toolname invokes an MILP solver to exhaustively search a region of the loss landscape around the weight assignments of $D$'s final layer parameters with the goal of {\it tunnelling through and escaping the local minima}. The process is repeated until desired accuracy is achieved. In our experiments, we find that \toolname not only scales well to additional data and very large model sizes, but also outperforms all other competing methods in terms of rates of convergence and data efficiency. For regression tasks, \toolname produced models that, on average, had 31.5\% lower MSE in 48\% less time, and for classification tasks on MNIST and CIFAR10, \toolname was able to achieve the highest accuracy over all competing methods, using only 50\% of the training data that GD baselines required. 
\end{abstract}

\section{Introduction}
Over the last few years, considerable amount of research has gone into algorithms for training Deep Neural Networks (DNNs), and yet, Gradient Descent (GD) and its variants remain the dominant approach for DNN training~\cite{ruder2016overview}. The primary reason for this state of affairs is that GD-based training methods can easily handle a large variety of DNN architectures and are highly scalable in training very large DNN, achieving high accuracy with relatively little computational effort. 

Having said that, despite their incredible success, GD-based methods~\footnote{While a variety of GD methods are available today, we focus on methods that offer the best accuracy, are most scalable, and the most widely used as of this writing.} do suffer from a few significant weaknesses. First, GD and variants fundamentally lack the ability to distinguish between local and global minima, and hence may get stuck in local minima resulting in sub-optimal performance, generalization. Second, there are scenarios where GD and variants suffer from poor data efficiency, i.e., the amount of data needed to get reasonable accuracy can be very high. Finally, in recent years researchers have been able to show that DNNs suffer security, trust and robustness issues, e.g., adversarial attacks~\cite{papernot2016limitations}, and that training DNNs to be adherent to certain constraints is highly desirable~\cite{verma2019compliance}. Unfortunately, GD and its variants can neither provide any guarantees, nor can they straightforwardly handle highly non-differentiable constraints that typically arise in the context of security and reliability specifications. 

All these weaknesses suggest that there is considerable room for improvement, and there is an urgent need for researching new classes of DNN training algorithms. Recognizing the above-mentioned issues with GD and its variants, researchers have proposed Mixed Integer Linear Programming (MILP) solver-based training methods~\cite{icarte2019training}, among others. Such methods have the advantage that they can guarantee optimality, can alert users to infeasible problems, and handle highly non-differentiable constraints such as the ones that arise in security specifications that can potentially be added to the set of optimization constraints~\cite{gupte2013solving}. Unfortunately, solver-based methods suffer from significant problems of over-fitting to training data and very poor scalability vis-a-vis the size of the network being trained.

While there have been attempts to augment GD-based methods with optimizers (e.g., Adam) and learning rate scheduling techniques to overcome the oft-repeated problem of getting stuck in local minima, they do suffer from being heuristic in nature, i.e., they do not provide any gurantees that they have reached a global minima. Perhaps more importantly from a practical point of view, such additional optimizers also suffer from relatively poorer data efficiency.

To address these issues, we provide a new hybrid training algorithm for DNNs, called \toolname, based on a combination of GD and an MILP solver (specifically we use the state-of-the-art Gurobi MILP solver~\cite{pedroso2011optimization}). Given a DNN $D$ and a training data set $S$ as input, \toolname initially invokes GD to train $D$ using $S$ until it gets stuck in a local minima (this can be detected using a variety of methods), at which point \toolname then invokes an MILP optimization solver to exhaustively search a region of the loss landscape around the {\bf current} weight assignments to {\it tunnel through and escape the local minima}. The GD and solver methods are invoked alternatively until an appropriate level of accuracy is achieved. When comparing \toolname against multiple GD baselines on a suite of regression and classification tasks, we find that \toolname not only scales well to additional data and model sizes, but also outperforms all other competing methods in terms of rates of convergence and data efficiency.

\begin{figure}[t]
\centering
\includegraphics[width=\linewidth]{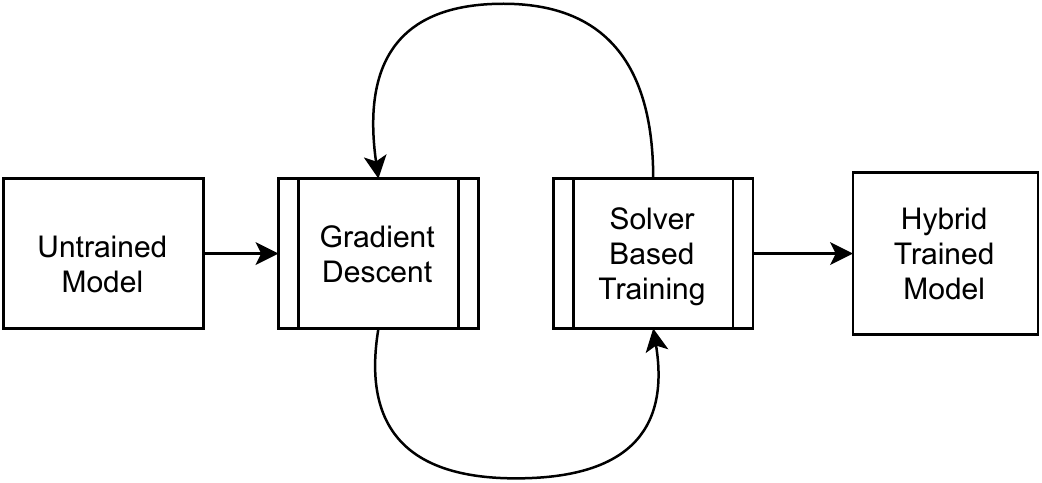}
\caption{Hybrid \toolname Architecture}
\label{fig:arch}
\end{figure}

\subsection{Key Contributions.}
\begin{enumerate}
    \item The \toolname Algorithm - a novel hybrid training algorithm that iteratively calls GD and MILP solver in a way that it is able to escape local minima by "tunneling" through them. In order to accomplish this, we had to come up with a novel formulation of DNN Training as an MILP instance that solves the severe overfitting problem of previous MILP formulations and enables its usage with real-valued DNNs. The \toolname algorithm is highly scalable in terms of being able to train very large DNN models, very general in terms of the DNN architectures it can handle, and data/resource efficient relative to competing methods~\footnote{Code available at: https://dhananjayashok.github.io/Hybrid-Solver-NN-Training/}.
    
    \item {\bf Extensive Experimental Evaluations:} We perform an comprehensive experimental evaluation of our algorithm against four state-of-the-art baselines, namely, Stochastic Gradient Descent (SGD), SGD with Learning Rate Scheduling (LRS), Adam Optimzation, and Adam Optimization with LRS, on a set of regression and classification tasks. 
    
    \begin{itemize}
        \item On a suite of regression equations, we show that \toolname can produce models with, on average, 31.5\% lower MSE in 48\% less time compared to state-of-the-art competing methods.
    
        \item On a set of standard classification datasets - MNIST and CIFAR10,  we show that \toolname is able to achieve the highest accuracy against all competing methods, using only 50\% the training data than competing GD baselines required for the same.
    \end{itemize}
\end{enumerate}

\begin{algorithm}[tb]
\caption{\bf The \toolname Algorithm}
\label{alg:tool}
\textbf{Input}: Untrained DNN, Training Data, Validation Data\\
\textbf{Parameter}: Desired Loss, MaxIter\\
\textbf{Output}: Trained DNN
\begin{algorithmic}[1] 
\STATE i := 0\;
\WHILE{Validation Loss is Decreasing}
\STATE Train DNN using Gradient Descent
\STATE Measure Validation Loss
\ENDWHILE
\STATE Convert final layer to an MILP Instance\;
\STATE Solve MILP Instance \;
\STATE Map Solution of MILP Instance back to NN parameters 
\STATE i++\;
\IF {Validation Loss $>$ Desired Loss and i $<$ MaxIter}
\STATE Go to Line 2
\ELSE
\STATE \textbf{return} trained DNN
\ENDIF
\end{algorithmic}
\end{algorithm}

\section{\toolname: The Architecture of Gradient Descent + Solver DNN Training Method}

In this section, we detail the steps outlined in the architecture diagram in Figure~\ref{fig:arch} and Algorithm~\ref{alg:tool} detailed above. The first step of \toolname is to train the network with GD alone, as one would for any DNN [lines 1-3], until the plateauing of validation loss, indicative of a local minimum, is observed [line 1]. At which point, \toolname halts the GD training and proceeds to the second step, namely, solver-based training. The value of using GD to train networks is well known, namely, scalability to very large networks and the ability of GD-based methods in obtaining low loss in many settings. 

In the solver phase [lines 5-7], the \toolname algorithm takes the partially trained network and focuses on "Fine Tuning" the final layer using an MILP solver. In this step, \toolname first converts the problem of training the final layer of the neural network to an MILP instance using a specialized formulation [line 5] (discussed in greater detail in Subsection~\ref{Formulation} below). The idea here is to search in a region around the values assigned by GD to the network's final layer weights and biases, such that the resultant assignment found by the solver has even lower loss than the one found by GD alone in step 2 (assuming such a lower loss point exists). If no lower loss point is found, \toolname stops training and returns the trained DNN [line 11]. 

The MILP instance thus formulated is solved by an MILP solver [line 6] (specifically, the Gurobi solver), and the solution is then mapped back to the network weights and biases [line 7]. We refer to this process as {\it final layer fine-tuning}. The termination condition for the training loop is a check that ascertains whether the desired accuracy has been achieved or further improvements to the weights and biases are possible. If yes, then the loop continues, else it terminates [lines 8-12].

\subsection{Motivation and Advantages of a Hybrid Solver+GD Training}

Since the MILP solver is only involved in fine tuning, and does not train the entire network end-to-end, it is completely invariant to the architecture and size of the Neural Network until the final layer. This hybrid approach makes \toolname much more scalable than prior methods for training neural networks using solvers alone~\cite{icarte2019training}. Further, the specific design choices we have made enables our method \toolname to be very general, i.e., handle a variety of architectures, since a large variety of DNN architectures can be symbolically modelled as MILP problems. At the same time, our method \toolname retains the ability to be highly influential on the final prediction strength of the DNN, as shown in our experiments.

The idea of fine tuning the final layer(s) alone is not new, and other methods have been proposed with dramatic and highly consequential impact on network performance~\cite{howard2018universal,pan2009survey}. To the extent we know, our work is unique in that we use a MILP solver for final layer fine tuning.

\begin{figure}
    \includegraphics[width=\linewidth]{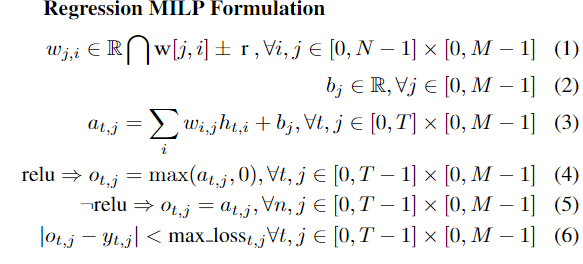}
    \caption{Regression Formulation}
    \label{fig:RegressionFormulation}
\end{figure}

\section{Escaping Local Minima via \toolname vs. Competing Methods} 

\noindent{\bf Escaping Local Minima via Solvers:} As has been noted, GD performs well until it gets stuck in a local minima, and limited options exist to escape local minima. By contrast, the solver performs exhaustive search in a large space around the weights and biases assigned by GD in the previous iteration, and thus may discover a new point with lower loss. By assigning this new point (i.e., new weights and biases) to the DNN, it may be able to escape the local minima by tunnelling through it and with the help of an additional iteration of GD more effectively than only using GD.  

\noindent{\bf Adam Optimizer and LRS:} Current alternatives for handling local minima include using momentum based optimization methods (Adam~\cite{kingma2014adam}, RMSProp~\cite{kurbiel2017training} etc.) and LRS~\cite{li2019exponential}. Momentum based optimization methods struggle in several instances due to the fact that they too use gradient information to decide the size of steps taken in training. While they are very good at finding the right step size to take, they are not so useful if the direction towards the better solution is currently unknown or not discoverable given local gradient information. Learning Rate Scheduling methods, while simple and efficient , are often highly dependent on hyper-parameter tuning and thus can be unreliable. 

The strength of the \toolname Algorithm is that it can be used alongside all of these above methods, and offers another route to escaping local minima if they perform poorly in particular settings. In the rest of this work, we focus on showing that our hybrid method is uniquely useful in efficient training on widely useful metrics of efficiency and generalization.

\begin{figure}
    \includegraphics[width=\linewidth]{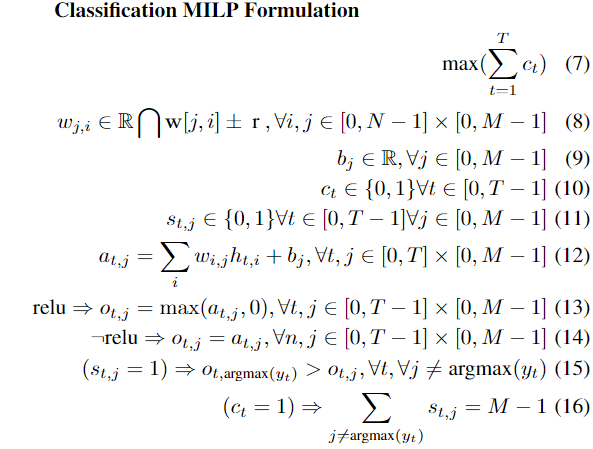}
    \caption{Classification Formulation}
    \label{fig:ClassificationFormulation}
\end{figure}

\begin{center}
 \begin{table*}[th]
 \centering
\begin{tabular}{|c|cc|cc|cc|}
\hline
Equation ID: & \multicolumn{2}{c|}{GD(10)}          & \multicolumn{2}{c|}{GD(20)}          & \multicolumn{2}{c|}{\textbf{\toolname}}                   \\ \hline
Equation     & \multicolumn{1}{c|}{MSE}    & Time   & \multicolumn{1}{c|}{MSE}    & Time   & \multicolumn{1}{c|}{\textbf{MSE}}      & \textbf{Time}   \\ \hline
Identity     & \multicolumn{1}{c|}{0.579}  & 0.0353 & \multicolumn{1}{c|}{0.2412} & 0.084  & \multicolumn{1}{c|}{\textbf{0.109}}    & \textbf{0.043}  \\
Affine       & \multicolumn{1}{c|}{16.467} & 0.0278 & \multicolumn{1}{c|}{8.075}  & 0.071  & \multicolumn{1}{c|}{\textbf{7.2095}}   & \textbf{0.0321} \\
Polynomial   & \multicolumn{1}{c|}{93.86}  & 0.0324 & \multicolumn{1}{c|}{20.805} & 0.076  & \multicolumn{1}{c|}{\textbf{12.07024}} & \textbf{0.0387} \\
Formula      & \multicolumn{1}{c|}{10.44}  & 0.0361 & \multicolumn{1}{c|}{4.208}  & 0.0864 & \multicolumn{1}{c|}{\textbf{3.32117}}  & \textbf{0.0452} \\ \hline
\end{tabular}
\caption{{\bf Regression Experiment Results:} Values for best GD baseline (SGD with LRS) GD benchmarked at halfway point (10) and final epoch (20). Results show GDSolver after 10 epochs of GD outperforms 20 epochs of GD for both MSE and Time}
\label{table:regressionTable}
\end{table*}
\end{center}

\section{The MILP Formulation} \label{Formulation}

Key to the success of our \toolname method is a symbolic formulation of the final layer of a Neural Network as an instance of the MILP problem, and then mapping the solution obtained by invoking an MILP solver back to the parameters of the final layer. Put differently, we convert the final layer of the DNN into a mathematical formula as described below. For our base formulation we use a variation of that used in \cite{icarte2019training}, with significant improvements as discussed here. The full formulation is presented in Figures~[\ref{fig:RegressionFormulation}] and  [\ref{fig:ClassificationFormulation}].

Similar to previous work in symbolic formulations of DNNs~\cite{bunel2017unified,cheng2017maximum}, we restrict our system to using only linear piece-wise or the soft-max activation functions for the final layer. In the formulation shown in the Figures~[\ref{fig:RegressionFormulation}] and [\ref{fig:ClassificationFormulation}], we restrict ourselves to the constraint for the ReLU activation function as this is most commonly used one. Having said that, our formulation can easily handle any linear piece-wise activation. 

\subsection{Setup and Definitions}
Let $f$ denote a partially trained Neural Network and $L$ its final layer, and $\{X, Y\}$ denote a dataset with $T$ datapoints. Let N denote the input dimension of the final layer $L$ and M be the dimension of the output. Then, $L:\mathbb{R}^{N\times 1}\to\mathbb{R}^{M\times 1}:$ is a map that can be written out using a weight matrix $\mathbf{w} \in {\mathbb{R}}^{M \times N}$, a bias vector $b \in \mathbb{R}^{M \times 1}$, input $h \in \mathbb{R}^{N \times 1}$ and activation function $\sigma$ as follows: $L(h) = \sigma(\mathbf{w}h+b)$. We can express $f$ as: $f = L\circ f'$ where $f'$ is all the previous layers of the DNN other than the final layer $L$. Then, the goal of DNN training is to learn the mapping $L\circ f'(X) = y \Leftrightarrow \sigma(\mathbf{w}h+b) = y$, where $h = f'(X)$.

\subsection{Base Formulation for Regression and Classification}
When converting this final layer $L$ to an MILP instance, all the weights and biases of the final layer are represented as variables $w_{i, j}$ and $b_{j}$. Constraints (1,2) in Figure~\ref{fig:RegressionFormulation} (respectively, constraints (8,9) in Figure~\ref{fig:ClassificationFormulation}) are box constraints that bound the region around the value assigned to the variables $w_{i, j}$ and $b_{j}$ that the solver is required to search.

Constraints (3, 4, 5, 12, 13, 14) as given in Figures[~\ref{fig:RegressionFormulation}, \ref{fig:ClassificationFormulation}], essentially encode the architecture of the neural network. More precisely, for every data point $(x_t, y_t)$ we compute $h_t = f'(x_t)$, where $h_t \in H$ is the input to the final layer. Then, constraint (3) (respectively, constraint (12)) encodes the input to the activation function as a linear combination of $h_t$ and the parameters of the final layer, while the constraints (4, 5, 13, 14) encode the activation with ReLU. Finally, each training data point also has constraints to relate the output of the neural network to the intended target label/ value $y_t$. The encoding of the target label depends on whether the problem is one of regression or classification. 

\subsection{Formulation of DNN Output}
\noindent{\bf Regression Output:} In the regression formulation, constraint (6) bounds the L1 distance of the output and target by a constant value max\_loss$_{t, j}$. In practice we set max\_loss$_{t, j} = L1(o, y)_{t, j}$, i.e., the L1 loss of that data point using the current weight assignment of the network. This ensures that if a solution is found by the solver, then it has a better L1 loss on the training dataset than the current assignment given by the previous GD step of the \toolname algorithm. 

\noindent{\bf Classification Output:} The output dimension of classification models is typically equal to the number of classes that could be predicted, and the prediction of the neural network is the class which corresponds to the highest output neuron value in the final layer for a given data point. With this in mind, constraints (15, 16) encode whether a given data point is correctly classified by the DNN, i.e., the variable $c_t$ is 1 iff datapoint $t$ is correctly classified, $\sum_{t}c_t$ is hence a measure of the total accuracy of the network. We set this accuracy as the maximization objective of the solver constraint (7), setting a lower bound on the accuracy as the current accuracy of the model given by the prior GD step of \toolname tool ensures that any solution found will have a better accuracy on the training set than the current assignment.  

\subsection{Mapping Solutions to the DNN}
Given this formulation, it is fairly straightforward to convert the final layer of the given network to a valid MILP problem and query a solver for satisfying assignments for the weights and biases. If a feasible solution is found we simply assign $W[i, j] = w_{i, j} \forall i, j$ and $b[j] = b_{j} \forall j$. 

\subsection{Discussion on Formulation}
There are several key differences between our formulation and the original used in \cite{icarte2019training}, that enable us to train networks faster and with less over-fitting. 

\textbf{Local Neighbourhood Restrictions:} The first is in how we define the weight and bias variables in constraints (1, 2, 8, 9) - we ensure that these variables can only be set to an interval around the current weight and bias assignments given by the prior GD step of the \toolname algorithm.  This restricts the space of assignments that the solver has to search over, vastly improving its scalability. It has the additional advantage to preventing over-fitting, as as new solutions cannot be too "different" from the current assignment. Finally, in practice, this restriction does not seem to impede the solver from tunnelling through the local minima.

\textbf{Regression Flexibility: }Instead of regression constraint (6) and classification (7, 15, 16), previous formulations relate the output of the neural network to the intended target label with the constraint $o_{j, t} = y_{j, t} \forall j, \forall t$~\cite{icarte2019training,thorbjarnarson2020training}. While this is more straightforward, it has significant flaws. This forces the network to search for assignments that perfectly regress every single training point which is highly likely to cause over-fitting. Our alternative of constraint (6) acknowledges that perfect accuracy on the training set is undesirable, striking a better balance by simply requiring a lower loss than the current assignment. This observation was absolutely key to the success of our formulation and DNN training tool.

\textbf{Classification Flexibility: }The problem with previous formulations of the constraints that relate the network output to labels is even more pronounced in the classification domain, where the constraint not only requires the solver to achieve perfect accuracy on the training set, but also requires the output vectors to match~\cite{icarte2019training,thorbjarnarson2020training}. The output vector for classification problems are often vectors of binary indicators of class membership for each class. For example, given a 3 class classification problem target vector $y_{\text{3 class} \in \mathbb{R}^{T\times 3}}$ where $y_{i, \text{3 class}} = [1, 0, 0]$ would mean that the $i$th data point is of the first class. In the vast majority of cases Neural Networks output vectors do not attempt to predict the exact 0-1 value, but rather predict un-normalized probabilities, such that the final prediction is the class with the highest output value in the predicted vector. Thus if pred$_{i} = [0.75, 0.2, 0.05]$ or pred$_{i} = [5, 1, 3]$ the DNN has correctly predicted the output label, but the previous MILP formulations would consider all of these to be incorrect as they do not match the exact vector $[1, 0, 0]$. Our formulation has constraints (7, 15, 16) which allow the model to fall short of perfect training accuracy and permits the DNN to predict un-normalized probabilities in its final layer.

\textbf{Generalization of previous formulations: }The strength of our novel formulation is that it is a generalization of the ones that came before, i.e., the previous formulations are a special case of ours that is obtained by setting the max\_loss and minimum accuracy parameters to 0 and 1 respectively.

\section{Experimental Evaluation}

\noindent{\bf Experimental Setup:} For all our experiments, we compare our hybrid method against the four GD baselines of SGD, SGD with LRS, Adam Optimization and Adam with LRS. The experiments were run on a system with the following specs: 18.04.2-Ubuntu with Intel(R) Core(TM) i7-10750H CPU @ 2.60GHz. Models were created and trained with standard PyTorch implementations of GD baselines, the datasets for MNIST and CIFAR10 were the standard datasets provided by PyTorch~\cite{paszke2019pytorch}. The MILP solver used was the python interface of the Gurobi solver - Gurobipy \cite{pedroso2011optimization}. 

\subsection{Experiment 1: Regression} \label{regressionExperiments}
In this experiment, our goal was to ascertain whether the \toolname algorithm and tool achieves faster convergence with greater data and resource efficiency than GD baselines. In order to make the comparison as fair as possible, we use regression datasets (namely, identity, affine, polynomial of degree 4, and trignometric and exponential formula. See Appendix for more details) that we know the baseline models can accurately predict with a low loss. 

The experiments were performed as follows: we vary the number of epochs $e$ (from 1 to 20) and for each GD baseline note the testing loss and time taken after $e$ epochs. We then compare this to the testing loss and time taken to complete $e$ epochs of SGD and a single solver sweep of the final layer. We expect to see a strict increase in time taken, as the hybrid method does all the iterations that the baselines do and an additional step, however if the improvement in loss is sufficiently large, then it would justify the additional time cost. This also allows us to quantify how many additional epochs of GD would have been required to achieve equivalent loss. 

\noindent{\bf Analysis of Results:} The results for the first experiment can be seen in Table~\ref{table:regressionTable}. For brevity the results shown are for the best GD baseline (SGD with LR Scheduling) at the median epoch and maximum epochs used - 10 and 20. (Figures showing complete results for all epochs can be found in the Appendix). The table compares the GD baseline after 10 and 20 epochs to \toolname after 10 epochs. The results suggest that the hybrid solver method is very useful in quickening the rate of convergence of loss - For each of these datasets, after 10 epochs, the hybrid method outperforms the other baselines with respect to generalization, and in most cases only after more than 20 epochs would the baselines catch up to the hybrid solvers generalization loss. The time taken by \toolname is greater than the baselines at 10 epochs, but significantly less than the time taken for 20 iterations, which is how long the baselines take to achieve a comparable loss - \toolname produced models with, on average, 31.5\% lower MSE in 48\% less time. These two observations put together motivated the conclusion that the hybrid solver method is an efficient and valuable method to quicken the rate of convergence, outperforming classic GD approaches. 

\subsection{Experiment 2: Classification}
In this experiment we ascertain whether our Solver+GD hybrid approach consistently performs better than GD baselines in terms of generalization to a test set as the amount of training data is varied. 

We perform the experiment by varying the number of training datapoints $n$ and epochs $e$. For each pair of these variables $(n, e)$ we train the baseline GD methods with $n$ points for $e$ epochs. For comparison, using the \toolname algorithm we train for a maximum of $\frac{e}{2}$ epochs of SGD ($n$ datapoints), stopping early and calling the solver if we detect a loss plateau - we do this process 2 times. This 2 Loop \toolname method uses a {\bf maximum} of $e$ epochs in its GD steps, hence making sure that any improved performance is not a consequence of simply more computation. When calling the solver, we do not give it the entire training dataset, but rather a single batch (32 datapoints) of datapoints which are incorrectly classified by the current assignment given by GD. We compare the GD baselines with the 2 Loop \toolname on test accuracy that measures generalization. We perform the above experiment on two well-known datasets - MNIST, CIFAR10. 

\begin{figure}[th]
  \includegraphics[scale=0.5]{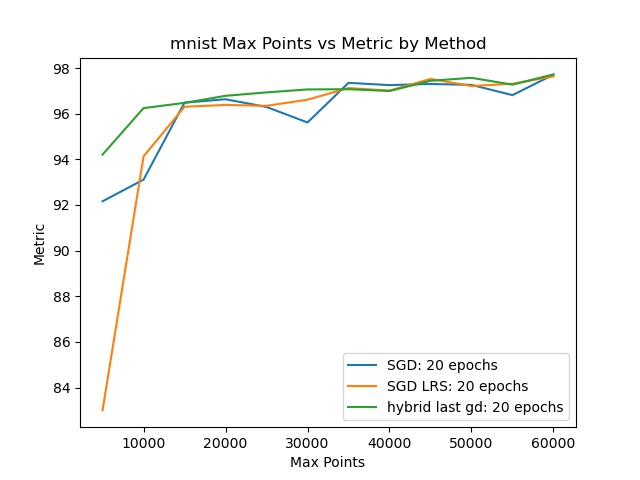}
  \caption{\toolname (green, hybrid last gd) achieves 96\% accuracy with half the datapoints that it takes the best GD baseline to}
  \label{fig:mnist}
\end{figure}

\begin{figure}[th]
  \includegraphics[scale=0.5]{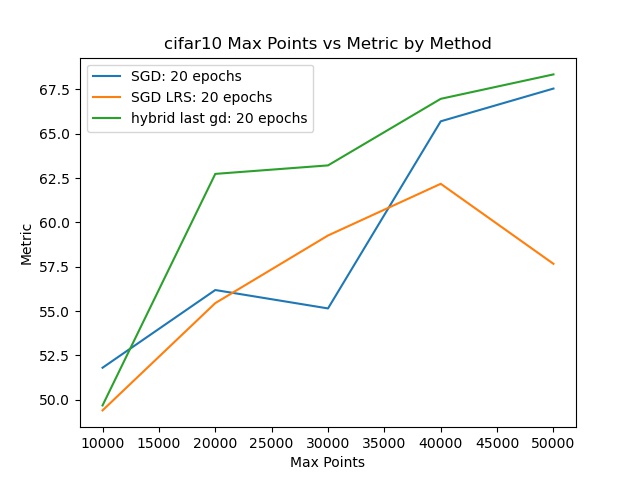}
  \caption{\toolname (green, hybrid last gd) is consistently better than both baselines, especially with less data}
  \label{fig:cifar10}
\end{figure}

\vspace{0.2cm}
\noindent{\bf Analysis of Results:} The results for the second experiment are shown in Figures [\ref{fig:mnist}] and [\ref{fig:cifar10}]. Results show that on average the hybrid method outperforms all GD baselines in terms of testing accuracy. The trend lines give deeper insight into the advantage that the hybrid method provides. It shows that consistently and significantly the hybrid method performs much better when fewer data points are used, i.e., the hybrid method is more data efficient. For both the datasets we can observe a trend where as training data increases the baseline methods eventually achieve similar performance to our method - \toolname on MNIST and CIFAR10 achieves better accuracy than GD baselines with only 50\% of the training data. This phenomenon of solver-based training methods performing better when data is scarce has been noted before. For example, in \cite{icarte2019training} they showed that for binary neural networks, models trained by solvers with limited data were vastly superior to GD methods. The results are also consistent with the idea that the solver sweep in \toolname tool helps with tunnelling through the local minimum that the GD baselines struggle with, and thus reaching higher accuracy with fewer data points than the GD baselines.

\noindent{\bf Limitations:} The \toolname makes a fundamental tradeoff, namely, in order to work with real valued DNNs it can only perform solver sweeps on the final layer of the network, for otherwise the resultant optimization would be a non-linear optimization problem and hence outside the scope of MILP solvers. \toolname also currently assumes that the DNNs it takes as input use a feed-forward densely connected final layer. This assumption is mostly true for regression and classification problems, however does not hold for most specialized networks like Image generating GANs etc.  

\section{Related Work}

\noindent{\bf Symbolic Formulation of DNNs:} There is growing literature in the interpretation of Neural networks symbolically as MILP, SAT, or SMT problems~\cite{tjeng2017evaluating,zhang2018efficient,bunel2017unified}. Almost all of this work is aimed towards verification of pretrained neural networks via DNN verification solvers (see VNN-LIB Initiative for more details). This has important implications as it is a fundamentally different formulation to the one we use in our tool, wherein the variables in the symbolic formulation for verification are input data points, while the variables in the context of training are network parameters.

\noindent{\bf DNN Training via Solvers:} Training neural networks using solvers has mainly been studied in the binary and integer settings. Narodytska et al.~\cite{narodytska2019search} studies converting Binary Neural Networks to SAT problems and studied which architectures were more "SAT" friendly so they may be solved efficiently. Icarte et al.~\cite{icarte2019training} established the first MILP formulation for Binary Neural Networks for training purposes. They attempt to deal with problems of over-fitting using regularizing objective functions and show that MILP solvers outperform GD as the training algorithm of choice when there is a sparsity of data points. Thorbjarnarson et al.~\cite{thorbjarnarson2020training} uses this same formulation and attempts to extend the analysis to integer valued neural networks. However both these methods suffer from scalability, over-fitting and cannot be used on real-valued networks~\cite{icarte2019training}.

\section{Conclusions and Future Work}
We present \toolname, a hybrid solver and GD training algorithm for DNNs, that consistently outperforms 4 state-of-the-art GD methods on several regression and classification tasks in terms of higher accuracy with greater data and resource efficiency. Further, to the best of our knowledge, ours is the first solver-based method that can scale to real-world sized DNNs. MILP Solvers and GD excel in different settings and ways. Our method \toolname leverages the advantages of each method, giving rise to an algorithm that combines the best of both worlds and is better than its individual parts on measures such as accuracy and data efficiency. By using solvers to {\it tunnel through and escape the local minima}, we tackle one of the most important and difficult problems that GD methods often encounter. This gives \toolname the ability to scale well to additional data and model sizes, but also outperforms pure GD methods in terms of rates of convergence and data efficiency. In the future, we plan to extend our work to handling highly non-linear constraints, since other classes of solvers, such as SMT solvers, are capable of handling such non-linearity. Further, one of our goals is to train DNNs in a manner that ensures (probabilistic) adherence to security or reliability constraints. It is unclear how a purely GD-based method can be used to provide guaranteed adherence to such specifications. By contrast, we believe that solver-based hybrid training methods could enable us to train DNNs in a manner that ensures (probabilistic) adherence to logical specifications.

\bibliography{library}

\end{document}